\def\year{2020}\relax
\documentclass[letterpaper]{article} % DO NOT CHANGE THIS
\usepackage{aaai20}  % DO NOT CHANGE THIS
\usepackage{times}  % DO NOT CHANGE THIS
\usepackage{helvet} % DO NOT CHANGE THIS
\usepackage{courier}  % DO NOT CHANGE THIS
\usepackage[hyphens]{url}  % DO NOT CHANGE THIS
\usepackage{color}
\usepackage{graphicx} % DO NOT CHANGE THIS
\usepackage{graphicx}  % DO NOT CHANGE THIS
\usepackage{mathtools}
\usepackage{algorithm}
\usepackage{algorithmic}
\usepackage{threeparttable} 
\usepackage{booktabs}
\usepackage{multirow}
\usepackage{lipsum}

\title{Patchy Image Structure Classification Using Multi-Orientation Region Transform}
\author{\Large \textbf{Xiaohan Yu\textsuperscript{\rm 1,2}, Yang Zhao\textsuperscript{\rm 1,2}, Yongsheng Gao\textsuperscript{\rm 1,\thanks{Corresponding authors}}, Shengwu Xiong\textsuperscript{\rm 2,*}, Xiaohui Yuan\textsuperscript{\rm 2}}\\ 
\textsuperscript{\rm 1}School of Engineering and Built Environment, Griffith University, Australia\\ %If you have multiple authors and multiple affiliations
\textsuperscript{\rm 2}School of Computer Science and Technology, Wuhan University of Technology, China\\
\{xiaohan.yu2, yang.zhao4\}@griffithuni.edu.au, yongsheng.gao@griffith.edu.au, 
\{xiongsw, yuanxiaohui\}@whut.edu.cn % email address must be in roman text type, not monospace or sans serif
}

\begin{document}

\maketitle

%%%%%%%%% ABSTRACT
\begin{abstract}
   Exterior contour and interior structure are both vital features for classifying objects. However, most of the existing methods consider exterior contour feature and internal structure feature separately, and thus fail to function when classifying patchy image structures that have similar contours and flexible structures. To address above limitations, this paper proposes a novel Multi-Orientation Region Transform (MORT), which can effectively characterize both contour and structure features simultaneously, for patchy image structure classification. MORT is performed over multiple orientation regions at multiple scales to effectively integrate patchy features, and thus enables a better description of the shape in a coarse-to-fine manner. Moreover, the proposed MORT can be extended to combine with the deep convolutional neural network techniques, for further enhancement of classification accuracy. Very encouraging experimental results on the challenging ultra-fine-grained cultivar recognition task, insect wing recognition task, and large variation butterfly recognition task are obtained, which demonstrate the effectiveness and superiority of the proposed MORT over the state-of-the-art methods in classifying patchy image structures. Our code and three patchy image structure datasets are available at: https://github.com/XiaohanYu-GU/MReT2019.   
\end{abstract}

%%%%%%%%% BODY TEXT
\section{Introduction}

\begin{figure}
\begin{center}
  \includegraphics[width=1\linewidth]{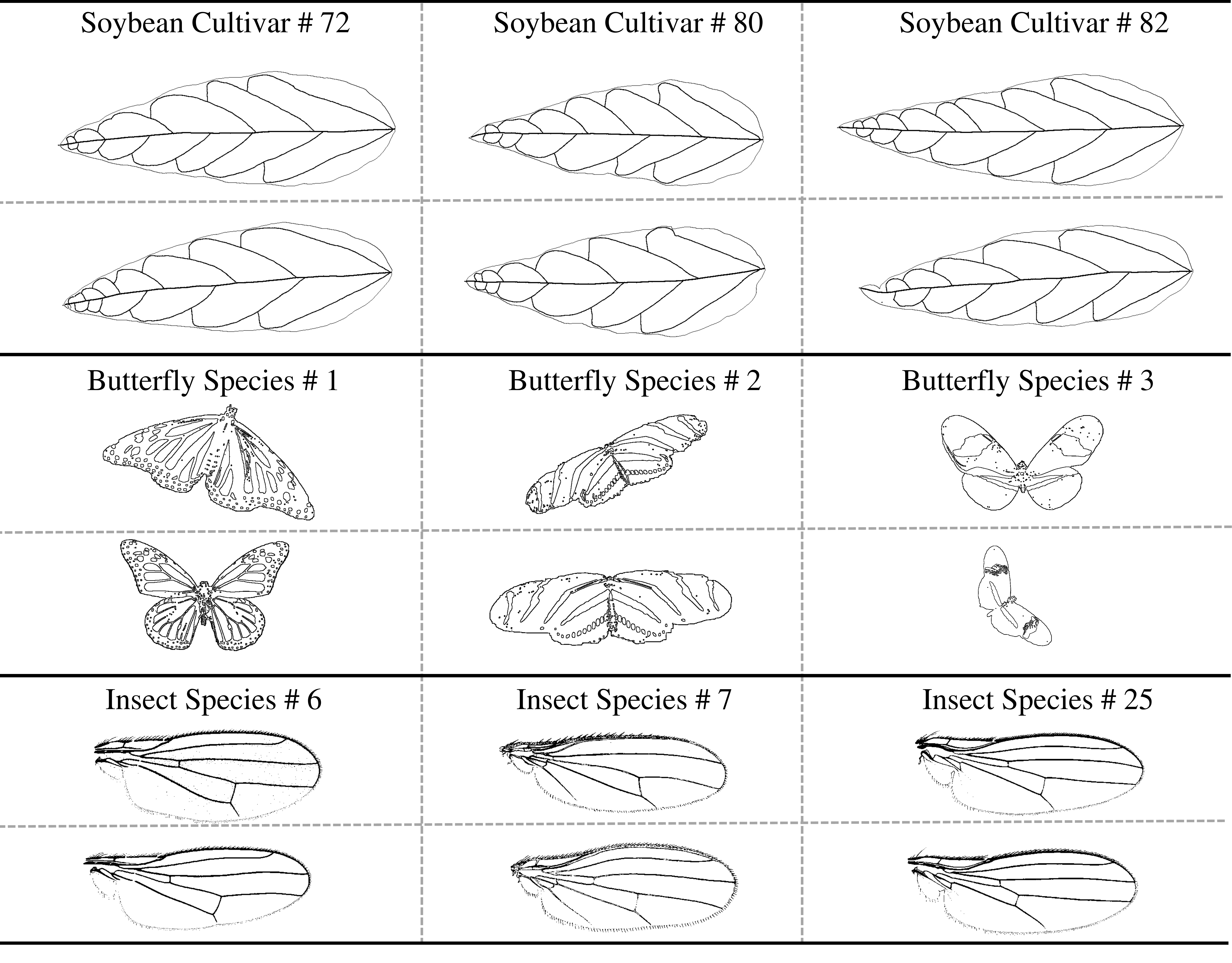}
\end{center}
  \caption{An example of illustrating the small inter-class differences (compare horizontally) in comparison to their intra-class variations (compare vertically) in patchy image structure classification.}
  \label{fig_1}
%   \vspace{-0.5cm}
\end{figure}

Patchy image structure classification is a fundamental yet significant research topic in computer vision and artificial intelligence (AI) research communities. Recent years have witnessed great progress of patchy image structure classification in applications highly related to AI agriculture and smart farming, such as plant branch structure classification \cite{8}, leaf vein classification \cite{9}, and insect wing vein classification \cite{41}. Patchy image structures characterize topological information of both exterior contours and interior structures from a target shape, which are important for shape modeling and classification. An example of illustrating the patchy image structures is shown in Fig. \ref{fig_1}.  Unfortunately, classifying such patchy image structures remains as an open problem, due to the fact that target shapes from various categories may have highly similar contours and flexible interior structures. 

In the past decades, many researchers have devoted their efforts to addressing this challenging problem. Existing approaches can be broadly classified into two categories: (1) handcrafted feature based methods and (2) deep convolutional neural network (ConvNets) techniques. Most approaches in the first category focus on developing effective feature representations of the shape contour \cite{11,20,7,28,25,4}. Although these methods achieved promising performance in general shape classification tasks, they may fail to function when the contour of objects among different classes are highly similar, such as the patchy image structures shown in Fig. 1.     

Alternatively, one may consider applying ConvNets techniques, which are currently driving advances in image classification tasks, to the patchy image structure classification. The most powerful and practical ConvNets methods are the so-called backbone networks, such as Alexnet \cite{52}, VGG-16 \cite{21}, and ResNet-50 \cite{34}, which have been proven to be very effective in large-scale image classification tasks. More recently, approaches \cite{fu2017look,Zheng_2019_CVPR,Yang2018Learning,chen2019destruction} focusing on fine-grained classification may also contribute to the challenging patchy image structure classification. However, a major limitation is that deep learning techniques rely heavily on the amount of the training samples, $i.e.$, they may fail to perform when limited training samples are provided \cite{37}.

In this paper, we address the patchy image structure classification problem by proposing a novel multi-orientation region transform method that can simultaneously characterize exterior contour, flexible interior structure, and inter-relationship between them. The contributions of our work are summarized as follows: (1) A Multi-Orientation Region Transform (MORT) is proposed to provide unified discriminative description of the flexible interior connection structure, shape of the exterior contour, and their inter-relationship in the patchy distance map of a patchy image structure with arbitrary orientation. (2) The proposed MORT is theoretically and experimentally proved to be rotation and translation invariant, which is important for effective and practical shape description and matching. (3) The proposed MORT can be extended to combine with ConvNets based features for further performance enhancement. (4) The encouraging experimental results demonstrate the effectiveness and efficiency of the proposed MORT over the state-of-the-art benchmarks on patchy image structure classification.

%------------------------------------------------------------------------
\section{Related work}
Existing methods for patchy image classification can be coarsely categorized into handcrafted feature based methods and deep convolutional neural network (ConvNets) techniques. A common strategy of handcrafted feature based methods is to present contour-based feature representations, which has yielded desirable performance for shape classification on various publicly available shape databases \cite{7,11,25,28}. Another line of work focuses on deformation-based shape analysis and classification. A recent progress on deformation-based shape analysis \cite{20} enables simple physical interpretations of the resulting deformations. However, their proposed similarity metric is based on the differential quantities which are highly sensitive to noise and local perturbations. To overcome the sensitivity to local perturbations, a deformation based curved shape representation (DBCSR) \cite{4} is proposed based on the relative transformation matrices between neighboring points, such that shapes are represented as elements of matrix Lie group. However, these methods may fail to function when classifying shapes with similar contours in ultra-fine granularity (e.g. leaf images from the same species but different cultivars). 

Recently, ConvNets methods have been successfully applied in various visual tasks, thus may also contribute to the challenging patchy image structure classification. Most existing ConvNets methods are designed based on the so-called backbone networks, such as Alexnet \cite{52}, VGG-16 \cite{21}, and ResNet-50 \cite{34}. Given the effectiveness and practicality of these backbone networks in image classification tasks, they are naturally regarded as the first options for exploring new classification tasks \cite{wu2019ip102}. 

Another promising alternative is to apply the fine-grained classification techniques, which are particular capable of classifying similar shapes. \citeauthor{lin2017improved} (\citeyear{lin2017improved}) introduced statistics normalization methods to improve an architecture that can capture second-order statistics of convolutional features in a translationally invariant manner. \citeauthor{li2018towards} (\citeyear{li2018towards}) proposed to utilize the second-order information to improve the fine-grained classification performance. 
Some researchers propose to first locate the discriminative regions and then classifies based on these regions \cite{huang2016part}. Such methods, however, require additional bounding box annotations on targeted regions. Another technique focuses on localizing discriminative regions by attention mechanism in an unsupervised manner, without requiring extra annotations. Unfortunately, these methods \cite{zheng2017learning,fu2017look,Zheng_2019_CVPR} require additional network structure (e.g., attention mechanism), leading to extra computation overhead for both training and inference stages. To that end, \citeauthor{Yang2018Learning} (\citeyear{Yang2018Learning}) proposed a method that is weak supervised without requiring the annotations of parts or key areas. \citeauthor{chen2019destruction} (\citeyear{chen2019destruction}) introduced a very effective Destruction and Construction (DCL) method for fine-grained classification. In their work, the input images are partitioned into local regions and then shuffled by a region confusion mechanism, leading to a classification network focusing on discriminative regions for spotting the subtle differences. Given a large number of training data, ConvNets based methods can yield desirable performance on classification tasks. However, the ConvNets based methods with limited training data may dramatically overfit the training data \cite{37}.

Aforementioned limitations motivate us to develop an effective and efficient method, without requiring large training sets for the challenging patchy image structure classification. Our work has the same assumption with the state-of-the-art handcrafted feature based methods \cite{3,11,7,28,25,20,4}, $i.e.$, the contours of shapes are assumed to be provided.

\section{Multi-orientation region transform method}

In this section, a novel Multi-Orientation Region Trans-form (MORT) method is proposed to encode not only global shape features but also local structures within a shape for finer-level structure pattern analysis. To better capture the interior structure, shape of the exterior contour, and their inter-relationship, a patchy distance map (PDM) is first developed. The proposed MORT (embedded with the PDM) and its discrete form is then described to construct the final feature descriptor. Finally, we present the rotation, translation and scale invariance analysis of the proposed feature descriptor, as well as the similarity measure using the transform coefficient matrices.

\subsection{Patchy distance map}
In set theory, when all sets under consideration are subsets of a given set $U$, the absolute complement of $A$ is defined as the set of elements in $U$ but not in $A$. Based on above notions, a patchy shape is defined to be composed of two subsets, interior set and its absolute complement. Specifically, given a shape $S$ with patchy structure, the patches are categorized into two types: (1) interior patches $I$, defined as closed sub-regions inside the shape when the contour is filtered out (e.g. closed pattern patches in a butterfly image); and (2) complementary patches $C$, defined as the absolute complement of interior patch $C=S \backslash I$. 

\begin{figure}
\begin{center}
  \includegraphics[width=1\linewidth]{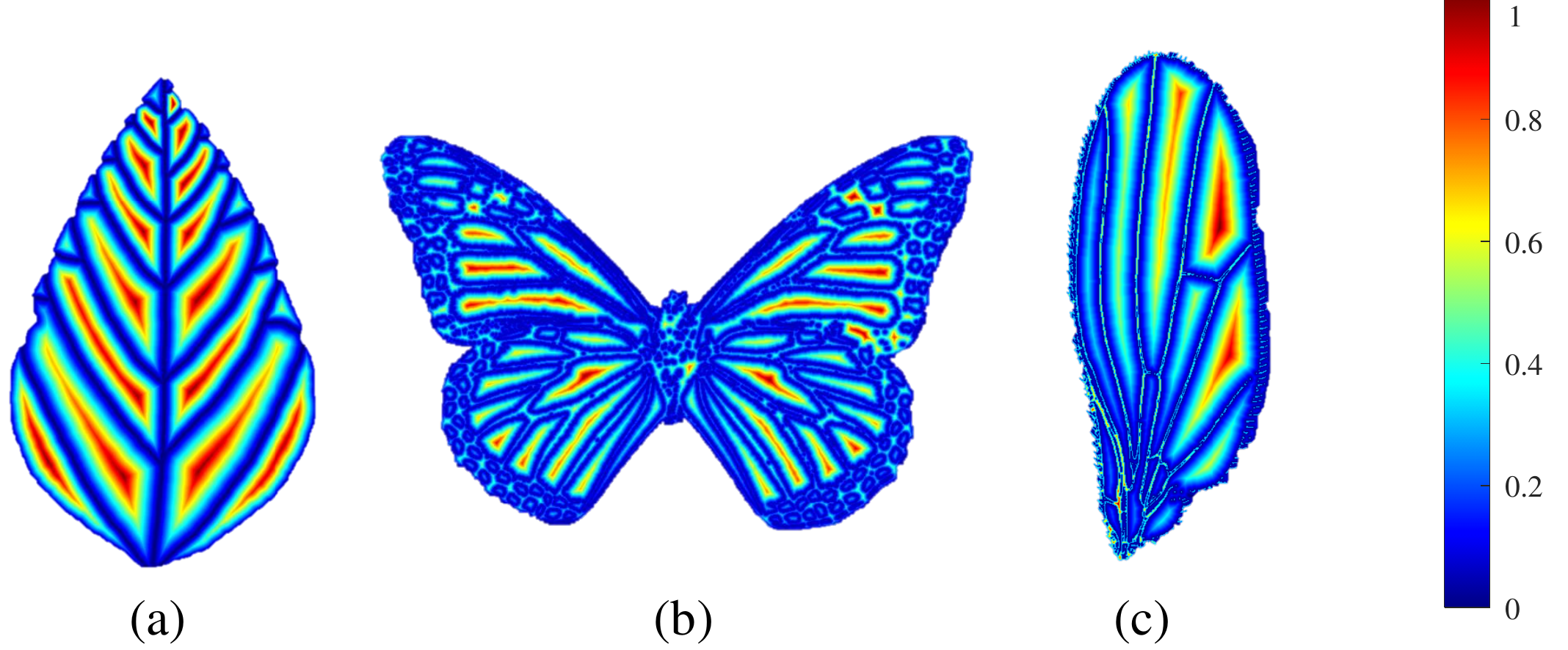}
\end{center}
  \caption{An example of illustrating distance maps of patchy image structures. The patchy structure distance map of (a) a soybean leaf image, (b) a butterfly image, and (c) an insect wing image. }
  \label{fig_dm}
%   \vspace{-0.2cm}
\end{figure}

In order to integrate both geometrical and topological features of the patchy shape, we propose a patchy distance map (PDM) to enable comprehensive skeleton-based feature description, in which each patch is encoded with distance transform. Specifically, we apply Euclidean Distance Transform \cite{12} to each patch, such that each pixel is assigned with a value computed by the distance between the pixel and the nearest patch boundary. By normalizing the distance map in each patch, both small patch and large patch can provide equally detailed description based on the PDM. An example of visualizing PDM is given in Fig. \ref{fig_dm}. For the convenience of description, we denote $f^I (x,y)$ as the interior patchy distance map function, and $f^C (x,y)$ as the complementary patch distance map function.

\subsection{Multi-orientation region transform}
\subsubsection{Region integral on a given contour point.}
\begin{figure*}[t]
\centering
 \includegraphics[width=1\linewidth]{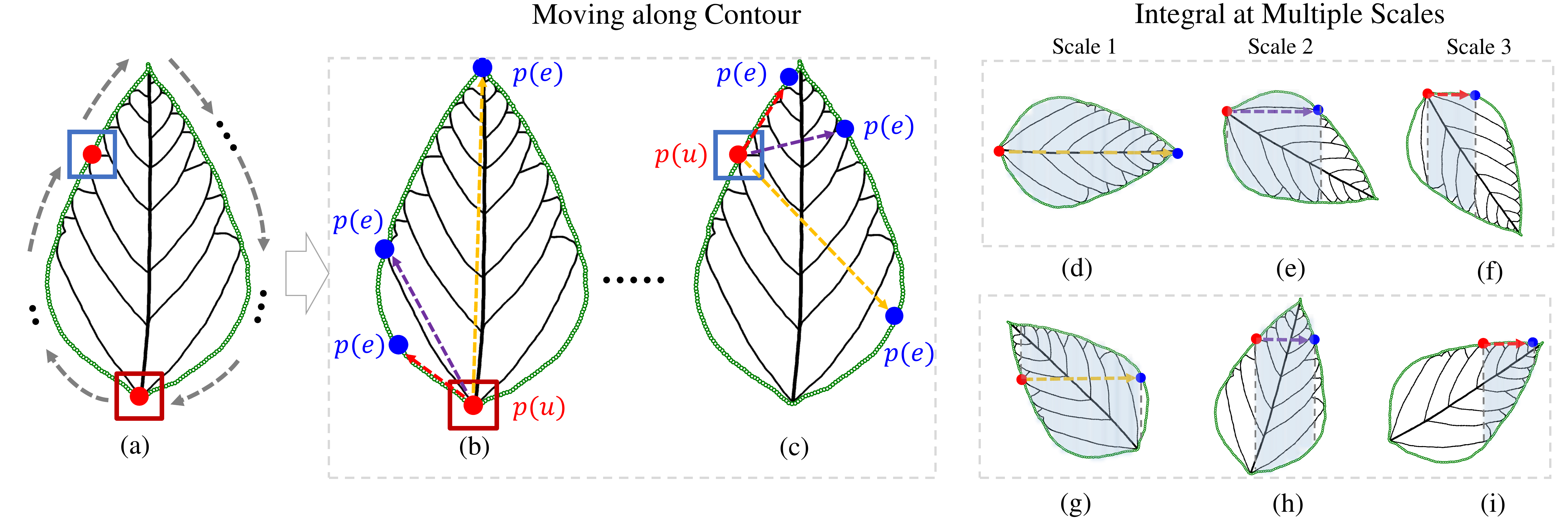}
\caption{An example of illustrating the proposed Multi-Orientation Region Transform on a leaf image. (a) Input image with two example locations of a moving $p(u)$ point. (b) and (c): Base lines (indicating the orientations of the region integral) for scales 1, 2 and 3 (shown in yellow, purple, and red respectively) for the two example locations of $p(u)$ shown in (a). (d), (e) and (f) display the scanning ranges of the region integral for the $p(u)$ shown in (b) at scales 1, 2, and 3, respectively. (g), (h) and (i) display the scanning ranges of the region integral for the $p(u)$ shown in (c) at scales 1, 2, and 3, respectively.}
\label{fig_method}
% \vspace{-0.1cm}
\end{figure*}

Given a contour point $p(u)$ ($u$ ranges from 0 to 1), the arc length $l(t)$, from $p(u)$ to its end point $p(e)$ on the contour along clock-wise direction, is defined as $l(t)=N/2^t$, where $N$ is the perimeter of contour (or the number of contour points when the contour is sampled in discrete form), and $t$ is the scale index ($t=0,1,\dots,Q$ where $Q=log_2 N$). Denote $p(u)$ and $p(e)$ by coordinates $(x(u),y(u))$ and $(x(e),y(e))$, a region integral on $p(u)$ at scale $t$ is defined as:
\begin{equation}
\begin{aligned}
& R_f^z (t,\theta_u^t) 
= \int_{-\infty}^{+\infty}\int_{x(u)\cos\theta_u^t+y(u)\sin\theta_u^t}^{x(e)\cos\theta_u^t+y(e)\sin\theta_u^t}\int_{-\infty}^{+\infty}\int_{-\infty}^{+\infty} \\ & f^z (x,y) \delta{(x\cos\theta_u^t+y\sin\theta_u^t-\lambda_u^t,x\sin\theta_u^t+y\cos\theta_u^t-\rho_u^t)} \\ & dxdyd\lambda_u^t d\rho_u^t,
\end{aligned}
\label{eq_1}
\end{equation}
where $ \lambda^t_u$ is the vertical distance from the origin to the boundary line defined by: $ x\sin \theta^t_u+y\cos \theta^t_u=\rho^t_u$. $\rho^t_u$ is the vertical distance from the origin to the base line defined by: $ x\cos \theta^t_u+y\sin \theta^t_u=\lambda^t_u$. $\theta^t_u$ is the angle between the base line and y-axis.  $z \in \{I,C\}$ and $\delta(\phi,\chi)$ is the 2D Dirac delta function defined as
\begin{equation}
\delta(\phi,\chi)=\begin{cases}
1& \text{if $\phi = \chi = 0 $}\\
0& \text{otherwise}
\end{cases}.
\label{eq_2}
\end{equation}
Here, $\delta(\phi,\chi)$ in Eq. (\ref{eq_1}) ensures that only the points on the intersection of base line $ x\cos \theta^t_u+y\sin \theta^t_u=\lambda^t_u$ and boundary line $ x\sin \theta^t_u+y\cos \theta^t_u=\rho^t_u$ are counted for integral calculation. Given a point indexed by $u$ under scale $t$, variables $\theta^t_u$, $\lambda^t_u$ and $\rho^t_u$ together determine the integral region, denoted as $\Omega_f (t,\theta_u^t )$, in Eq. (\ref{eq_1}). As $p(u)$ moves along the contour, the region integral is performed over different $\Omega_f (t,\theta_u^t )$ defined by the scanning range and orientation of the boundary line at all possible scales (see Fig. \ref{fig_method}).

\subsubsection{MORT coefficient matrix.}

The MORT on a given patchy distance map function $f^z (x,y)$ at point $p(u)$ is defined as
\begin{equation}
MT_f^z(\theta_u) =  [R_f^z (t,\theta_u^0) ,\dots, R_f^z (log_2 N,\theta_u^Q)]^T
\label{eq_3}
\end{equation}
The MORT at a single point $p(u)$ generates a vector of $(Q+1)$ coefficients with each describing the region integral for one scale at point $p(u)$. An example of the proposed MORT on point $p(u)$ is illustrated in Fig. \ref{fig_method}. By moving $p(u)$ along the contour for a complete loop, which results in a varying $\theta\in[0,2\pi)$, the above vector grows into a MORT coefficient matrix of $(Q+1)$ by $N$ dimensions:
\begin{equation}
\boldsymbol{MT}_f^z  = \left[
\begin{matrix}
R_f^z (0,\theta_1^0)    & \cdots & R_f^z (0,\theta_N^0)      \\
 \vdots  & \ddots & \vdots \\
 R_f^z (Q,\theta_1^Q)     & \cdots & R_f^z (Q,\theta_N^Q)     \\
\end{matrix}
\right]. 
\label{eq_4}
\end{equation}
This matrix describes how the region integrals at different scales synchronously vary when the point $p(u)$ moves. When scale $t=1,2,\dots,$ $p(u)$ and $p(e)$ cut a half, a quarter, $\dots$, off the contour to steer the region integral at different orientations (see Fig. \ref{fig_method}(b)\&(c)). The greater the $t$ is, the smaller integral region is scanned, which provides finer descriptions on local details of the target (see Fig. \ref{fig_method}(d)-(i). When scale $t=0$, $p(u)$ and $p(e)$ become the same point after a complete loop, the region integral is performed over the whole region of the given image, making the first row elements in matrix $\boldsymbol{MT}_f^z$ share the same value. For example, the 2nd, 3rd, 4th rows in matrix $\boldsymbol{MT}_f^z$  are visualized in Fig. \ref{fig_method}(d), Fig. \ref{fig_method}(e), and Fig. \ref{fig_method}(f) when point $p(u)$ moves to the position shown in Fig. \ref{fig_method}(b). 

Let $\{(x_k,y_k ) | k=1,\dots,N_f\}$ be the subset of pixels enclosed in the outer contour of the object shape, where $N_f$ is the total number of pixels enclosed in the contour. The procedure of computing the discrete MORT coefficient matrix $\boldsymbol{MT}_f^z$  is presented in Algorithm 1, in which Eq. (\ref{eq_1}) becomes the sum of patchy distance map (PDM) values inside the moving integral region $\Omega_f (t,\theta_i^t )$.

As Steps 2, 3 and 4 take time $O(N)$, $O(log_2 N)$ and $O(N_f)$ respectively, the algorithm 1 has a computational complexity of $O(N_f N log_2 N)$.
\renewcommand{\algorithmicrequire}{ \textbf{Input:}} 
\renewcommand{\algorithmicensure}{ \textbf{Output:}} 
\begin{algorithm}[t]
\caption{Calculating discrete Multi-Orientation Region Transform}
\begin{algorithmic}[1]
\REQUIRE ~~\\
$f^I (x,y)$: interior patch distance map;\\
$f^C (x,y)$: complementary patch distance map;\\
N: number of sampling points on contour;\\
\ENSURE ~~\\
$\boldsymbol{MT}_f^I$: interior MORT coefficient matrix;\\
$\boldsymbol{MT}_f^C$: complementary MORT coefficient matrix;\\
% \hline ~~\\
\STATE Initialize the two MORT coefficient  matrices as zero matrices: $\boldsymbol{MT}_f^I= \boldsymbol{MT}_f^C = \boldsymbol{0}_{(log_2 N+1)\times N}$;
\FOR{$i=1$ to $N$}
\FOR{$t=0$ to $log_2 N$}
\FORALL {$k$ such that $(x_k,y_k) \in \Omega_f (t,\theta_i^t )$ } 
\STATE  $R_f^I (t,\theta_i^t )=R_f^I (t,\theta_i^t )+ f^I (x_k,y_k )$; 
\STATE  $R_f^C (t,\theta_i^t )=R_f^C (t,\theta_i^t )+ f^C (x_k,y_k )$; 
\ENDFOR
\ENDFOR
\STATE   $\boldsymbol{r}_i^I=[R_f^I (0,\theta_i^0 ),\dots,R_f^I (log_2 N,\theta_i^{log_2 N}) ]^T$;   
\STATE  $\boldsymbol{r}_i^C=[R_f^C (0,\theta_i^0 ),\dots,R_f^C (log_2 N,\theta_i^{log_2 N}) ]^T$;  
\ENDFOR
\STATE $\boldsymbol{MT}_f^I=[\boldsymbol{r}_1^I,..,\boldsymbol{r}_N^I]$;        \STATE $\boldsymbol{MT}_f^C=[\boldsymbol{r}_1^C,..,\boldsymbol{r}_N^C]$;       
\label{code:recentEnd}
\end{algorithmic}
\end{algorithm}

\subsubsection{Final MORT feature descriptor.}

By extracting features from various orientations and scales, the image region can be described in a coarse-to-fine manner to provide comprehensive description of the shape. More importantly, this allows encoding the inter-relationship of contour and interior context of a shape. 

When the initial location of the contour point, that steers the Multi-Orientation Region Transform (i.e., the red point in Fig. \ref{fig_method}, moves clockwise, the entire column of $\boldsymbol{r}_i^z$ in $\boldsymbol{MT}_f^z$  shifts to the right. The magnitudes of its 1D Fourier transform coefficients are calculated by
\begin{equation}
 \widetilde{R}^z (t,k)= (\frac{1}{N}) \left|\sum_{i=1}^N R_f^z (t,\theta_i^t) exp{(-\frac{2\pi jik}{N})}\right|,
 \label{eq_5}
\end{equation}
where $k=1,\dots,N$ and $t=0,\dots,log_2 N$. To make the generated feature descriptor robust to noise and compact, the lowest M order coefficients are used to describe the target, where $M\ll N$. And $\widetilde{R}^z (t,k)$ are used to construct the final feature descriptor as
\begin{equation}
\boldsymbol{\Psi}_f^z = \left[
\begin{matrix}
\widetilde{R}^z (0,1)    & \cdots & \widetilde{R}^z (0,M)      \\
 \vdots  & \ddots & \vdots \\
 \widetilde{R}^z (log_2 N,1)     & \cdots & \widetilde{R}^z (log_2 N,M)    \\
\end{matrix}
\right]. 
\label{eq_6}
\end{equation}

\subsection{Invariance analysis and similarity measure}

In this section, we present the rotation, translation and scale invariance analysis of the feature descriptor, which is important for effective and practical shape matching \cite{6}. 

\subsubsection{Invariance analysis.}

\textbf{Lemma 1}: feature descriptor $\boldsymbol{\Psi}_f^z$ is rotation invariant. It is not difficult to prove the rotation invariance. According to Eq. (\ref{eq_5}) and Eq. (\ref{eq_6}), every element in $\boldsymbol{\Psi}_f^z$ is invariant to the initial location of the contour point that steers the MORT, and thus enable $\boldsymbol{\Psi}_f^z$ to be invariant to rotation of the whole shape. 

\textbf{Lemma 2}: A translation of $f^z (x,y)$ by a vector $
\Vec{u}=(x_0,y_0)$ (i.e. $f^z (x-x_0,y-y_0 )=h^z (x,y))$ do not change any element in $\boldsymbol{MT}_f^z$ : $R_f^z (t,\theta_u^t )=R_h^z (t,\theta_u^t )$. This shows that every element in $\boldsymbol{MT}_f^z$ is translation invariant, and thus $\boldsymbol{\Psi}_f^z$ is translation invariant. 

\textbf{Lemma 3}: A scaling of $f^z (x,y)$ by a factor $\gamma$ (i.e. $f^z (\gamma x,\gamma y)=h^z (x,y))$ changes each element in $\boldsymbol{MT}_f^z$ by a factor $1/\gamma^2$ : $R_h^z (t,\theta_u^t )=$  $1/\gamma^2 R_f^z (t,\theta_u^t )$.  In summary, the feature descriptor $\boldsymbol{\Psi}_f^z$ is invariant to rotation and translation, and becomes scaling invariant if it is normalized by the area of the shape.

Given two sets of matrices $\boldsymbol{T}= \cup_k \boldsymbol{\Psi}^z_{(f(T)_\_k)} =\{ \boldsymbol{\Psi}^z_{(f(T)_\_1)} ,\dots, \boldsymbol{\Psi}^z_{(f(T)_\_K)} \}$ (refer to Eq. (\ref{eq_6})) and $\boldsymbol{M}= \cup_k \boldsymbol{\Psi}^z_{(f(M)_\_k)} =\{ \boldsymbol{\Psi}^z_{(f(M)_\_1)} ,\dots, \boldsymbol{\Psi}^z_{(f(M)_\_K)} \}$ representing the MORT coefficient matrices of the test and model patchy shapes respectively. Each matrix set is comprised of $K$ matrix pairs of $\boldsymbol{\Psi}^I_f$ and $\boldsymbol{\Psi}^C_f$ and $K$ is the number of types in each class. Note that $\boldsymbol{\Psi}^z_{(f(T)_\_k)}$ in $\boldsymbol{T}$ can only be matched against the matrix of the same type in $\boldsymbol{M}$, that is $\boldsymbol{\Psi}^z_{(f(M)_\_k)}$  with the same type index $k$.
The dissimilarity between two given samples can be measured using the fast $L_1$ Minkowski distance of the two matrices:
\begin{equation}
\begin{aligned}
    & dis(\boldsymbol{\Psi}^z_{(f(T)},\boldsymbol{\Psi}^z_{(f(M)})
    = \sum_{k=1}^K[|\boldsymbol{\Psi}^I_{(f(T)_\_k)}-\boldsymbol{\Psi}^I_{(f(M)_\_k)}| \\
    & +|\boldsymbol{\Psi}^C_{(f(T)_\_k)}-\boldsymbol{\Psi}^C_{(f(M)_\_k)}|].
\end{aligned}
\label{eq_7}
\end{equation}
It is worth noting that the proposed method (if needed) can be used as feature extractor together with a classifier, such as the classic support vector machine (SVM), for classification tasks \cite{17}.

\section{Experiments and discussions}

In this section, we first introduce three publicly available patchy image structure datasets for classification evaluation. We then present the experimental results including both classification accuracy and computational time of the proposed MORT together with state-of-the-art methods. Finally, we present a fused MORT by integrating MORT with ConvNets methods via feature-level fusion, and report evaluation results on the three patchy image structure datasets.

\subsection{Patchy image structure datasets}
\begin{figure}
\begin{center}
 \includegraphics[width=1\linewidth]{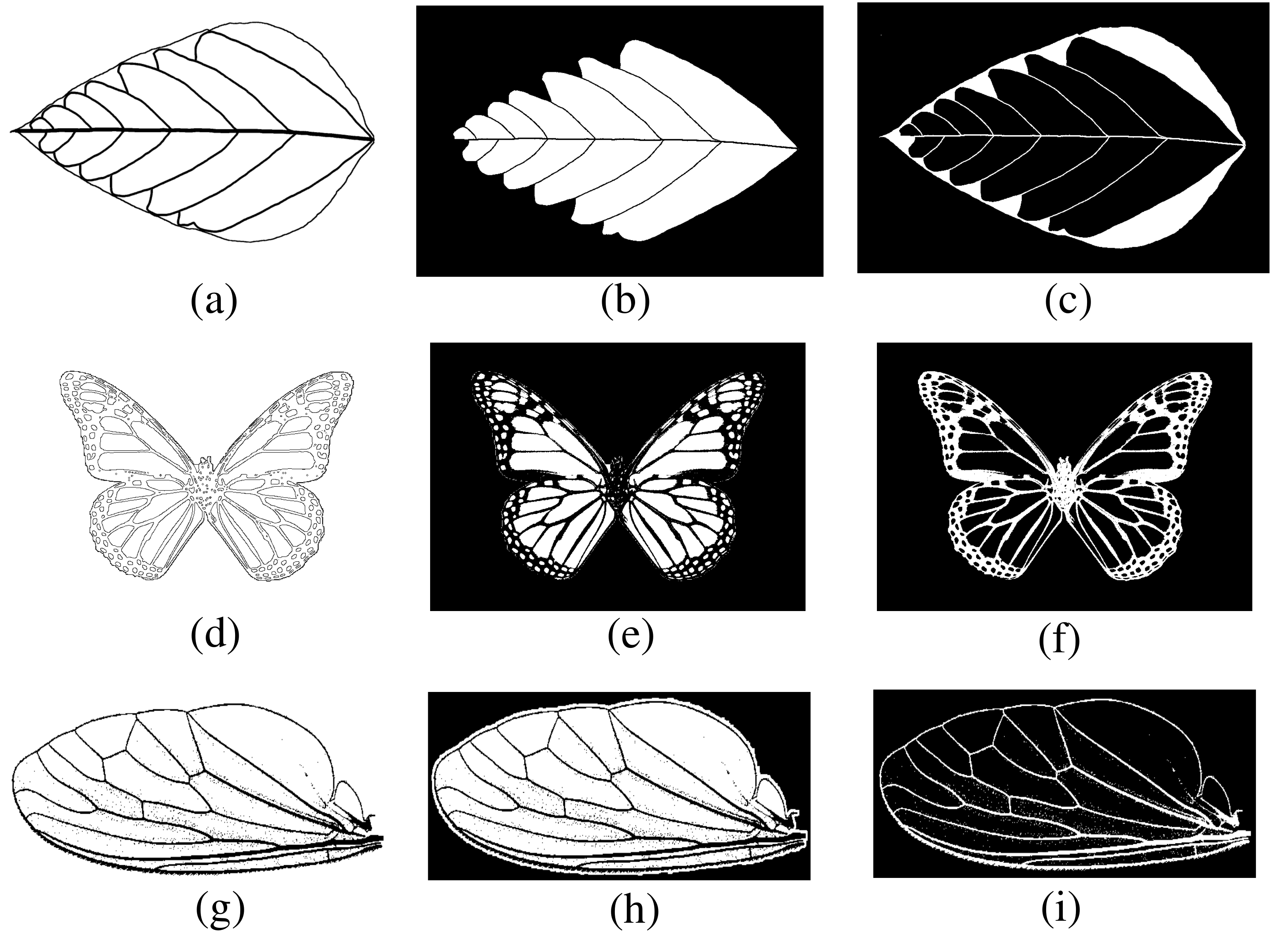}
\end{center}
\caption{An example of illustrating (a), (d), (g) patchy image structures and their corresponding (b), (e), (h) interior patches (white patches), and (c), (f), (i) complementary patches (white patches).}
\label{fig_patch}
\end{figure}
 \subsubsection{SoyCultivarVein dataset.} The SoyCultivarVein dataset is a publicly available dataset, which comprises 100 categories (cultivars) with 6 samples (leaf images) in each cultivar and thus has a total number of 100$\times$6 = 600 images  \cite{Yu_IEEEAccess2019}. The leaves in the SoyCultivarVein dataset are highly similar due to the fact that they all belong to the same species, making it a new and challenging dataset for the artificial intelligence and pattern analysis research community. 
\subsubsection{BtfPIS dataset.} The butterfly patchy image structure dataset (BtfPIS), is constructed by applying the canny edge detection \cite{33} to the binarized images of the first 50 images in each class from the public available Leeds butterfly dataset \cite{32}. There are 50$\times$10 = 500 images in the BtfPIS dataset. The BtfPIS dataset comprises large intra-class variations including rotation and scaling variations in each class.

\subsubsection{IwPIS dataset.} The insect wing patchy image structure dataset (IwPIS) \cite{38}, is adopted for evaluation, which comprises 25 classes of insect wings with 2 samples in each class. Although small, the IwPIS is very challenging for classification tasks due to the fact that all the samples are from the same order called Diptera. In other words, samples in IwPIS have very small inter-class variances that are difficult to be classified.

\begin{table}
\caption{Accuracy $\&$ matching time on SoyCultivarVein dataset (K=3)}
\begin{center}
\begin{tabular}{  l c l }
\hline\hline
\specialrule{0em}{2pt}{2pt}
\multirow{1}{*}{Algorithm} & Accuracy (\%) & Time (ms)\\ 
\specialrule{0em}{2pt}{2pt}
\hline
\specialrule{0em}{1pt}{1pt}
SC  & 37.58 & $2.38 \times 10^0$\\
SC+DP  & 36.90 & $1.17 \times 10^{-1}$\\
IDSC  & 49.07 & $2.26 \times 10^0$ \\
IDSC+DP  & 46.93 & $1.14 \times 10^{-1}$\\
SRV  & 37.09 & $6.89 \times 10^{2}$\\
DBCSR (opt) & 35.72 & $3.69 \times 10^{4}$\\
DBCSR (uni)  & 34.89 & $3.63 \times 10^{4}$\\
HSC  & 43.98 & $7.43 \times 10^{-3}$\\
MDM  & 39.39 & $3.93 \times 10^{-2}$\\
HF  & 40.46 & $1.02 \times 10^{1}$\\
\textbf{Proposed MORT} & \textbf {53.43} & \boldmath {$ 6.62 \times 10^{-2}$}\\
\hline
\hline
\end{tabular}
\end{center}
\label{tab_soy}
\end{table}

\subsection{Comparisons with shape-based benchmarks}
We compare the proposed MORT with ten state-of-the-art shape-based benchmarks. They are: (1) two versions of Shape Contexts, i.e., standard Shape Contexts (SC) and Shape Contexts with dynamic programming (SC-DP) \cite{3}, (2) Inner Distance Shape Contexts (IDSC) and Inner Distance Shape Contexts with dynamic programming (IDSC-DP) \cite{11}, (3) square-root velocity (SRV)  method \cite{20}, (4) deformation-based curved shape representation (DBCSR) with uniform sampling and optimal sampling \cite{4}, (5) Hierarchical String Cuts (HSC) \cite{25}, (6) Multiscale Distance Matrix (MDM) \cite{7}, and (7) Height Functions (HF) \cite{28}. The widely used Nearest Neighbor score (1NN) is employed for performance measurement, as used in the benchmark methods. 

Note that, the  MORT does not restrain how the interior and complementary patches shall be defined. In our experiments, for SoyCultivarVein dataset, the interior patches are defined as the patches enclosed only by vein points, while the complementary patches are those enclosed by a mixture of vein and contour points (see Figs. \ref{fig_patch}(a-c)). For BtfPIS dataset and IwPIS dataset, the interior patches are patches with bright pixels (i.e., their intensities are above or equal to the binarization threshold), while the complementary patches are those formed by dark pixels (i.e., their intensities are below the threshold), as illustrated in Figs. \ref{fig_patch}(d-f) and Figs. \ref{fig_patch}(g-i), respectively.
\subsubsection{Evaluation on SoyCultivarVein dataset (ultra-fine-grained).}
Table \ref{tab_soy} illustrates the average classification accuracies of MORT together with the state-of-the-art shape-based methods. We repeat the classification evaluation 1000 times by reselecting different three samples randomly to construct the model set and the remaining samples as the testing set. The average results are reported. Using the 1NN evaluation protocol, the proposed MORT achieves the highest average classification accuracy of 54.20\% (10.22\% higher than the HSC and 5.13\% higher than the IDSC), demonstrating its superiority in the ultra-fine-grained cultivar classification task. Table \ref{tab_soy} also lists the computational cost (the average time for each matching) of all the competing methods. The matching speed of MORT is among the most efficient methods (ranked the second), demonstrating the efficiency of the MORT method.

In order to verify the rotation invariance of the proposed method, we construct a rotated SoyCultivarVein dataset, by rotating each leaf image in the SoyCultivarVein dataset with a random angle (from $0^\circ$ to $360^\circ$). We compare the proposed method with two rotation invariant shape classification methods, HSC and MDM on the rotated SoyCultivarVein dataset. The experimental results (see Table \ref{tab_rotsoy} in comparison with the Table \ref{tab_soy}) confirm the theoretical analysis on rotation invariance of MORT.

\begin{table}
\caption{Accuracy on rotated SoyCultivarVein dataset}
\begin{center}
\begin{tabular}{  l c }
\hline\hline
\specialrule{0em}{2pt}{2pt}
\multirow{1}{*}{Algorithm} & Accuracy (\%) \\ 
\specialrule{0em}{2pt}{2pt}
\hline
\specialrule{0em}{1pt}{1pt}
HSC & 45.11 \\
MDM & 39.30 \\
\textbf{Proposed MORT} & \textbf {54.81} \\
\hline
\hline
\end{tabular}
\end{center}
\label{tab_rotsoy}
\end{table}

\begin{table}
\caption{Accuracy $\&$ matching time on BtfPIS dataset (K=1)}
\begin{center}
\begin{tabular}{  l c l }
\hline\hline
\specialrule{0em}{2pt}{2pt}
\multirow{1}{*}{Algorithm} & Accuracy (\%) & Time (ms)\\ 
\specialrule{0em}{2pt}{2pt}
\hline
% \multicolumn{4}{ c }{Inter-ocular Normalization} \\
% \hline
\specialrule{0em}{1pt}{1pt}
SC  & 42.57 & $3.00 \times 10^0$\\
SC+DP  & 41.93 & $1.08 \times 10^{-1}$\\
IDSC  & 57.45 & $2.10 \times 10^1$ \\
IDSC+DP  & 54.32 & $1.13 \times 10^{-1}$\\
SRV  & 60.49 & $1.14 \times 10^{3}$\\
DBCSR (opt) & 52.95 & $4.23 \times 10^{4}$\\
DBCSR (uni) & 58.53 & $3.80 \times 10^{4}$\\
HSC & 60.78 & $8.10 \times 10^{-3}$\\
MDM  & 54.84 & $5.00 \times 10^{-2}$\\
HF  & 20.64 & $8.50 \times 10^{1}$\\
\textbf{Proposed MORT} & \textbf {75.02} & \boldmath {$ 5.45 \times 10^{-2}$}\\
\hline
\hline
\end{tabular}
\end{center}
\label{tab_btf}
\end{table}

\subsubsection{Evaluation on BtfPIS dataset (large intra-class variation).}

Table \ref{tab_btf} illustrates the average classification accuracies of MORT together with the state-of-the-art shape-based methods. We repeat the classification evaluation 1000 times by reselecting different 25 samples randomly to construct the test set (and the remaining 25 samples are used as the model set) and their average results are reported. Using the 1NN evaluation protocol, MORT achieves the highest classification accuracy of 75.02\%, which are significantly higher than the state-of-the-art benchmarks (14.24\% higher than the HSC and 17.57\% higher than the IDSC). The superior performance of the proposed method over the state-of-the-art benchmarks demonstrates the robustness of MORT in classifying images with large intra-class variances. The matching speed of MORT on BtfPIS dataset is also among the most efficient methods (ranked the second).

\subsubsection{Evaluation on IwPIS dataset (small inter-class difference).}

Table \ref{tab_Iw} shows the average classification accuracies and average matching time of MORT and the state-of-the-art methods on the challenging IwPIS dataset. Using the same repeating strategy and 1NN evaluation (half images as model set and the remaining as testing set) as adopted in previous experiments, MORT achieves the highest classification accuracy of 44.79\%, which is higher than other 10 benchmarks (4.1\% higher than the SC and 13.87\% higher than the IDSC). The average matching time (for each matching) of MORT is $ 1.56 \times 10^{-2}$ ms, which is the lowest among all the benchmarks. The results show the superiority of MORT against state-of-the-art benchmarks in classifying patchy image structures with small inter-class variances.

\subsection{Comparisons with ConvNets methods}
\subsubsection{Fused MORT.}
Combining complementary feature representations may significantly improve the classification performance \cite{wang2015beyond}. However, a key question is that whether the proposed MORT and the state-of-the-art ConvNets methods can provide each other complementary features for further performance improvement. To that end, we propose a fused MORT by concatenating the original MORT feature matrices and feature vectors extracted from a state-of-the-art ConvNets method, DCL \cite{chen2019destruction}, and then investigate the comparative results of ConvNets methods and the fused MORT. To facilitate the comparison between the proposed method and ConvNets methods, we combine the fused MORT (as the feature encoder) with SVM (as the classifier) using the same protocol as adopted by \citeauthor{wu2019ip102} (\citeyear{wu2019ip102}). 
\subsubsection{Competitors.}
We compare the proposed MORT with the following state-of-the-art ConvNets methods: (1) Three state-of-the-art ConvNets models: Alexnet \cite{52}, VGG-16 \cite{21}, and ResNet-50 \cite{34}; and (2) three fine-grained state-of-the-art methods: improved B-CNN \cite{lin2017improved}, fast-MPN-COV \cite{li2018towards}, and DCL \cite{chen2019destruction}.   % B-CNN \cite{lin2015bilinear}, 
\subsubsection{Implementation details.}
All the models are implemented in Pytorch 1.0.0 and are pretrained on the ImageNet \cite{deng2009imagenet} and then fine-tuned on each dataset. 
In all experiments, the input images are resized to 440 $\times$ 440, and cropped to 384 $\times$ 384 randomly for training. Standard data augmentations are applied including random rotation within $\pm$15 degree and horizontal flip with 0.5 probability.  For fast-MPN-COV and improved B-CNN, the models are trained with the best setting in the implementation of fast-MPN-COV \cite{li2018towards}. For the remaining methods, the models are trained with the default settings in the implementation of DCL \cite{chen2019destruction}.

\begin{table}
\caption{Accuracy $\&$ matching time on IwPIS dataset (K=1)}
\begin{center}
\begin{tabular}{  l c l }
\hline\hline
\specialrule{0em}{2pt}{2pt}
\multirow{1}{*}{Algorithm} & Accuracy (\%) & Time (ms)\\ 
\specialrule{0em}{2pt}{2pt}
%Algorithm & Accuracy (\%) & Time (ms) \\
\hline
% \multicolumn{4}{ c }{Inter-ocular Normalization} \\
% \hline
\specialrule{0em}{1pt}{1pt}
SC & 40.78 & $3.14 \times 10^0$\\
SC+DP & 37.59 & $1.40 \times 10^{-1}$\\
IDSC  & 30.92 & $2.92 \times 10^{0}$ \\
IDSC+DP  & 34.90 & $1.36 \times 10^{-1}$\\
SRV & 36.40 & $8.44 \times 10^{2}$\\
DBCSR (opt) & 33.75 & $4.12 \times 10^{4}$\\
DBCSR (uni)  & 27.72 & $4.20 \times 10^{4}$\\
HSC  & 35.15 & $ 1.72 \times 10^{-2}$\\
MDM & 27.06 & $ 4.08 \times 10^{-2}$\\
HF  & 36.96 & $1.42 \times 10^{1}$\\
\textbf{Proposed MORT} & \textbf {44.79} & \boldmath {$ 1.56 \times 10^{-2}$}\\
\hline
\hline
\end{tabular}
\end{center}
\label{tab_Iw}
\end{table}

\subsubsection{Evaluation on patchy image structure datasets.}
For SoyCultivarVein, BtfPIS and IwPIS datasets, we select the first half images from each category as the training set, and the remaining images as the testing set. Table 
\ref{tab_CNN} lists the classification accuracies of all the competing ConvNets methods on the three datasets. The fused MORT achieves the best classification accuracy of 65\% and 98.00\% on SoyCultivarVein dataset and BtfPIS dataset, respectively. Nevertheless, the fused MORT ranks third in classification accuracy on IwPIS, with 6\% and 12\% lower than the improved B-CNN and fast-MPN-COV, respectively. A possible reason is that the fused MORT has much lower feature dimension (1$\times$2092) per image compared with improved B-CNN (1$\times$262144) and fast-MPN-COV (1$\times$32896). The overall results show that the fused MORT surpasses the original MORT in classification accuracy on all the three patchy image structure datasets, indicating that ConvNets methods and the proposed MORT can provide each other with complementary information. The superior results of the fused MORT over ConvNets methods on three datasets indicate that the proposed MORT can extract discriminative features even with very limited training samples, leading to the performance improvements of fused MORT over the sole use of DCL or original MORT.

\begin{table}
\caption{Accuracy on three patchy image structure datasets}
\begin{center}

\begin{tabular}{l c c c}
\hline
\hline
\specialrule{0em}{2pt}{2pt}
\multirow{2}{*}{Algorithm} & \multicolumn{3}{c}{Accuracy (\%)}\\  \cline{2-4}
\specialrule{0em}{2pt}{2pt}
 &SoyCultivarVein&BtfPIS&IwPIS \\
\hline
\specialrule{0em}{1pt}{1pt}
 Alexnet & 14  & 93.20  & 44 \\
 VGG-16  &  16 & 95.60  & 48 \\
ResNet-50  & 28  & 96.80  & 52 \\
% NTS-Net & \\
% B-CNN & 4.67 & 49.6 &  12 \\
% improved B-CNN & 51 & 49.60 &  12 \\
improved B-CNN & 63 & 73.60 & 74 \\
% CBP & \\
% fast-MPN-COV  & 26   & 77.2   &  64 \\
fast-MPN-COV  & 51   & 77.20   &  80 \\
DCL & 39  & 97.20  & 68 \\
    %   DCL(no branch) &  58 & 96  & 68 \\
     %   DCL(no branch white background) &  53 & 96  & 68 \\
 %      \textbf{MORT DCL} & \textbf {65} & \textbf{97.60} & \textbf{68}\\ %15
 %   \textbf{MORT ResNet-50} & \textbf {44} & \textbf{97.60} & \textbf{64}\\ %100 for Btf 50 for Leaf
              \textbf{fused MORT} & \textbf {65} & \textbf{98.00} & \textbf{68}\\ %50 for Btf 15 for leaf
%   \textbf{MORT,20} & \textbf {64} & \textbf{98} & \textbf{64}\\ 
%  \textbf{MORT,15} & \textbf {47} & \textbf{97.60} & \textbf{68}\\ 
%   \textbf{MORT,20} & \textbf {46} & \textbf{98} & \textbf{64}\\ 
%\textbf{MORT ResNet-50} & \textbf {} & \textbf{97.20} & \textbf{76}\\ %10
\hline
\hline
\end{tabular}
\end{center}
\label{tab_CNN}
\end{table}

\section{Conclusion}

In this paper, we presented a novel Multi-Orientation Region Transform (MORT) method, which is rotation, translation and scale invariant, for effective and efficient classification of patchy image structures. The proposed MORT can extract local structural features at various scales and orientations for comprehensive shape description. The encouraging experimental results on three patchy image structure databases demonstrate the effectiveness and efficiency of MORT for patchy image shape classification. Moreover, the results also indicate that the proposed MORT and ConvNets methods can provide each other with important complementary features to further improve the classification performance. 

\section{Acknowledgement}
This work was supported in part by the Australian Research Council under Discovery Grant DP180100958 and Linkage Grant LP170100326. 

\bibliography{aaai}
\bibliographystyle{aaai}

\end{document}